\documentclass[11pt,letterpaper]{article}

\usepackage[margin=1in]{geometry}
\usepackage[T1]{fontenc}
\usepackage[utf8]{inputenc}
\usepackage{graphicx}
\usepackage{booktabs}
\usepackage{longtable}
\usepackage{enumitem}
\usepackage{caption}
\usepackage{amsmath}
\usepackage[numbers,sort&compress]{natbib}
\usepackage{textcomp}
\usepackage{hyperref}

\hypersetup{colorlinks=true, linkcolor=blue, urlcolor=blue, citecolor=blue}

\title{From Document Silos to Process Intelligence: A Dual-Layer Knowledge
Graph for CMC Process Development}

\author{%
  Reza Amirmoshiri\textsuperscript{1},
  Faryad Sahneh\textsuperscript{2}, and
  Yasser Jangjou\textsuperscript{1}%
}
\date{%
  \textsuperscript{1}CMC Synthetics, Sanofi US\\
  \textsuperscript{2}Accelerator Data Science GBU, Sanofi US\\[0.5em]
  \textbf{Correspondence:}~\href{mailto:yasser.jangjou@sanofi.com}{yasser.jangjou@sanofi.com}%
}

\begin{document}
\maketitle


\begin{abstract}
Chemistry, Manufacturing and Controls (CMC) process development generates an enormous body of technical information across a multi-stage, knowledge-intensive continuum from drug discovery to commercial manufacturing. This knowledge is traditionally fragmented across functions and heterogeneous formats, causing traceability gaps and significant knowledge-management costs during technology transfer and regulatory filing. We present a modular agentic-AI platform that converts a heterogeneous corpus of process-development documents into a queryable, dual-layer knowledge graph. A base knowledge layer builds a lexical graph with a Document\,$\to$\,Section\,$\to$\,Chunk hierarchy through lossless ingestion of digital, scanned, handwritten, and multilingual documents, while an intelligence layer extracts ontology-aligned entities and bridges cross-document concepts through a provenance-anchored domain graph. LLM agents operate across both layers, selecting the retrieval path best suited to each question.

We evaluate the lexical layer with a novel three-tier protocol measuring the deployment-fidelity of a retrieval-augmented generation (RAG) system on proprietary data, demonstrated on 505 questions curated from 38 development reports of a Sanofi small-molecule program. Tier-1 multiple-choice accuracy of 95\% signals strong platform reliability; the stricter Tier-2 LLM-judge pass rate of 85\%, which degrades on comparative and corpus-wide questions, reveals a failure taxonomy that Tier-1 accuracy alone fails to capture. A router agent selects between layers according to question type. We anticipate this protocol will enable future designers of agentic platforms to assess their systems against nonpublic databases, and that graph-based architectures will see broader adoption in pharma as a means of transforming fragmented document repositories into structured process intelligence.

\end{abstract}

\section{Introduction}

\begin{figure}[!htbp]
  \centering
  \includegraphics[width=\linewidth]{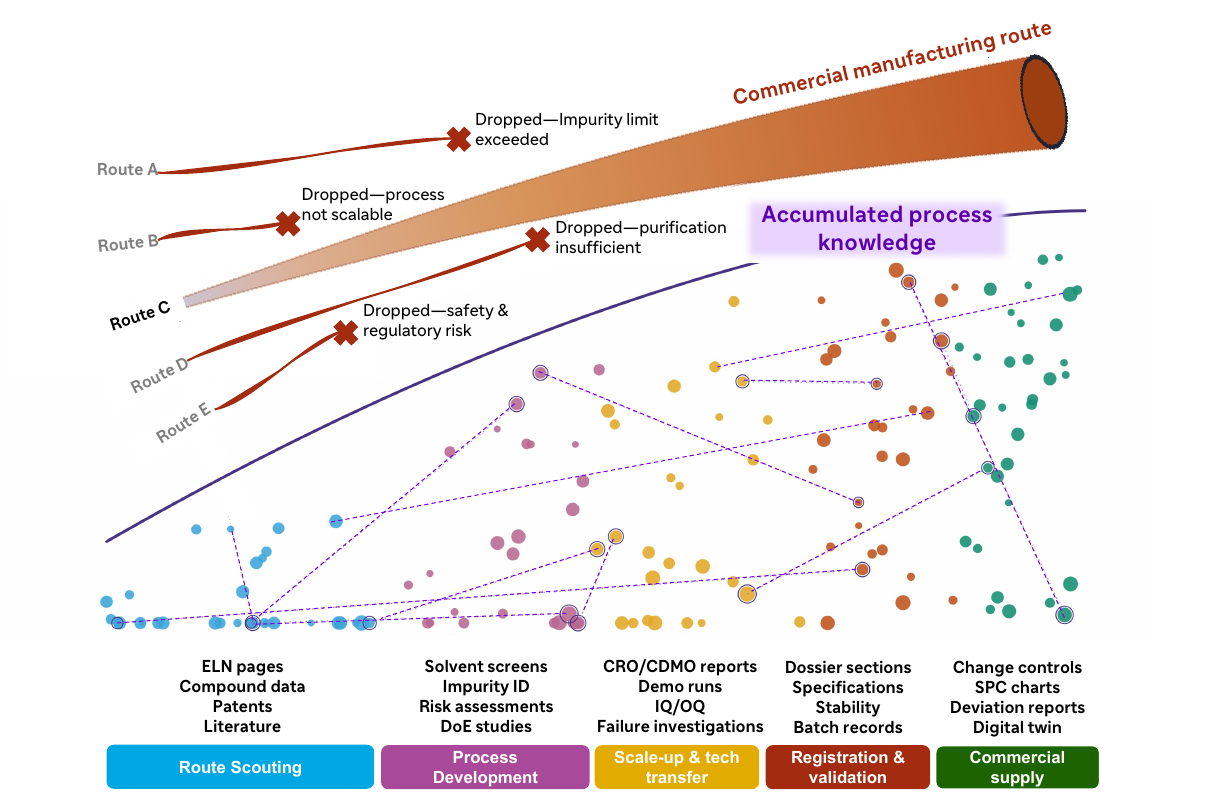}
  \caption{Knowledge accumulation and route convergence across a typical CMC
  process development timeline. Each phase draws on its own characteristic sources, named beneath it, and each node represents an information source. The most valuable
  insight lies not in a single record but in the relationships between them which span phases, scales, sites, and years. A platform based on knowledge graph transforms fragmented data into actionable knowledge. }
  \label{fig:figure1}
\end{figure}

Chemistry, Manufacturing, and Controls (CMC) serves as the critical bridge
between molecular discovery and therapeutic realization, encompassing the
operationally complex processes required to deliver a safe, effective and
manufacturable medicine \citep{fda_cmc_guidance, ich_q8, ich_q10, mustoe2025qbdd}. CMC development is a multi-stage, knowledge-intensive continuum extending from small-scale process scouting to robust, optimized, and scalable manufacturing for clinical and commercial supply. As illustrated in \autoref{fig:figure1}, an
enormous body of information accumulates during development while remaining fragmented across electronic laboratory notebooks (ELNs), technical reports, CMC dossiers, and batch records, in heterogeneous formats and languages. In this context, surveys report 10--15 years of development time and \$1--2 billion in R\&D costs for biopharmaceutical companies \citep{dimasi2016innovation, farid2020benchmarking,
Sertkaya2024, destro2022review}.

Although ongoing digitalization in the biopharmaceutical sector aims to reduce the tangible costs of development, the hidden costs of knowledge management and transfer remain largely unaddressed. Data fragmentation, coupled with manual document reviews, makes technology transfer and dossier preparation slow and error-prone, while version proliferation and expert-dependent interpretation introduce traceability gaps, siloed storage, and the risk of outright knowledge loss. Artificial intelligence (AI) systems have been shown to address these challenges in biomedical fields, spanning document ingestion, information retrieval, and knowledge synthesis \citep{bonner2022review, wu2025medical, gondi2026pharmasense, zhang2025comprehensive, li2022deepkg}. Custom AI chatbots, built on a retrieval-augmented generation (RAG) engine, have been commonly deployed to address user queries against an organization's proprietary documents. RAG conditions a large language model (LLM) on passages retrieved from a document store rather than on its parameters alone, thus improving the LLM's contextual awareness and addressing the limitation of the LLM's context window \citep{lewis2020rag, gondi2026pharmasense, taskiran2025knowledge}. RAG is particularly attractive for biopharmaceutical companies, where data privacy is a crucial concern \citep{buehler2024knowledgeextraction, wu2025medical, gondi2026pharmasense}. One of the few applications reported within CMC development is SUSIE (Schema-based Unsupervised Semantic Information Extraction), a question-answering system for drug discovery and development grounded in a knowledge graph \citep{mann2023susie, taskiran2025knowledge}.

Knowledge graphs (KGs) offer a fundamentally different representation from traditional data storage: knowledge is structured as a network of nodes
(entities) connected by edges (relationships) carrying labels and properties.
Rather than storing information in isolated tables or documents, a knowledge graph encodes entities and their relationships as an explicit, machine-readable, queryable network \citep{buehler2024knowledgeextraction}. LLMs have recently made
it practical to build KGs directly from unstructured technical text and to reason over them. This is achieved using GraphRAG, a RAG variant that augments flat-text retrieval with a graph so that the model reasons over explicit entities, relationships, and their neighborhoods rather than isolated chunks \citep{edge2024graphrag, khorshidi2025odke, xu2024ragkg}. LLM reasoning is influenced by graph type and structure: a lexical graph represents documents, sections, and chunks as its nodes, whereas a domain graph represents real-world or digital-world entities according to a defined ontology. Ontologies covering CMC topics have been reported in the literature \citep{allotrope2024afo, pistoia2024cmcontology, venkatasubramanian2006popeI, hailemariam2010popeII, viswanath2022ontology, remolona2017holmes}.

CMC development is inherently relational: the most valuable insights often lie not within a single information chunk but in the connections among data points across experiments, scales, sites, and time. This makes GraphRAG attractive for turning isolated data slices into a coherent, queryable whole. However, graph construction is costly and delicate and requires careful consideration of the endpoint application before construction begins. Question-answering based on keyword and semantic search can largely be addressed by a robust vector-RAG system coupled with a lexical graph---a precision layer. A CMC example is a process scientist searching for precise experimental conditions across hundreds of documents; the current wave of ``chat-with-your-documents'' demonstrations fits this category. However, corpus-wide anomaly detection (for instance, investigating the source and
fate of an impurity) instead requires cross-document and
cross-project retrieval over an ontological graph. To the authors' knowledge, no application within CMC development has addressed this endpoint variety with a modular architecture. Direct ingestion of source documents into a domain graph can miss the keyword-oriented questions that a process scientist most commonly asks. A second challenge is the quality of the domain graph, which depends on entity resolution, an essential concern in fields like process chemistry. For instance, acetonitrile, MeCN, ACN and $\mathrm{CH_3CN}$ all denote one canonical entity, although they might be written differently across technical reports; careful entity resolution must therefore precede domain graph construction. The third challenge is the absence of rigorous benchmarking specific to process development, as a persuasive chat transcript is not evidence of reliability. An end-to-end pipeline can fail at any stage of graph construction, knowledge retrieval, or inference.

The architectural idea of separating instance-level facts from domain-level
concepts in a dual-layer graph has recently been demonstrated outside
the pharmaceutical industry, in reservoir engineering \citep{li2025duallayer}, and the need for a CMC knowledge infrastructure grounded in an ontology has been argued at the policy level \citep{hussain2026nipte}. Our contribution is to realize this architecture within the context of CMC process development and, critically, to subject it to quantitative evaluation at a scale that reveals where it fails.

The current paper addresses the existing gaps with the following contributions:
\begin{itemize}
  \item An end-to-end modular pipeline for CMC process development that converts heterogeneous document sources into structured graph representations and answers questions through complementary LLM agents selected by question type. The graph architecture is dual-layer, combining lossless lexical ingestion of heterogeneous CMC documents (digital, scanned, handwritten, and multilingual) with an ontology-guided intelligence layer that bridges disconnected per-document graphs into a single cross-document, cross-project institutional graph. The pipeline also integrates document translation and scanned-document digitization, saving time and cost. We first quantify whether hybrid retrieval based on lexical graphs alone (without an applied ontology) can address a variety of questions asked by a CMC expert. We then illuminate the additional value of an ontology-driven domain graph anchored to the base lexical graph. 
  \item A three-tier benchmarking protocol and a question-design methodology that quantify reliability across increasing reasoning complexity, together with a failure taxonomy that pinpoints where and why retrieval breaks down. We emphasize the criticality of comprehensive benchmarking, which is largely unreported in previous work, not only within CMC but as part of the RAG platforms introduced in other fields. While competence benchmarks, such as ChemRAG-Bench \citep{zhong2025chemrag}, evaluate whether a system knows the context (chemistry, in that case), our protocol measures deployment fidelity, i.e., whether the system reliably retrieves and reasons over a specific institution's private documents/database. We believe that the benchmarking protocol introduced in this paper is applicable to future retrieval systems in various domains. 
  \item A direct analysis of lexical versus domain graphs with respect to the type of question, based on a sample of questions. We show that a hybrid system benefiting from both layers addresses a wider spectrum of questions than either alone; and that a router agent, which selects retrieval paths, improves both accuracy and traceability. This is made possible by making the domain graph \emph{provenance-anchored} to the lexical evidence: every canonical entity and relation resolves back to the exact source passages that justify it (\autoref{sec:intelligence-layer}).
\end{itemize}

Although the platform has been examined across multiple internal Sanofi
CMC Synthetics projects, this paper reports graph details and assessments for a single discontinued Sanofi small-molecule program. The paper proceeds as follows. First, we describe the platform architecture and its hybrid retrieval, together with the agentic workflow from document ingestion through graph construction to agent reasoning, followed by the evaluation approach and question design. Second, we report the dual-layer graph composition, followed by examining the lexical graph reliability and limits under three-tier benchmarking. Third, we demonstrate how anchoring a domain graph to the existing lexical layer, together with a router agent that selects the retrieval path according to the type of question, can address the limitations of the lexical graph. Finally, we discuss the key conclusions and outlook for future work.

\section{Platform Architecture and Methods}

\begin{figure}[!htbp]
  \centering
  \includegraphics[width=0.92\linewidth]{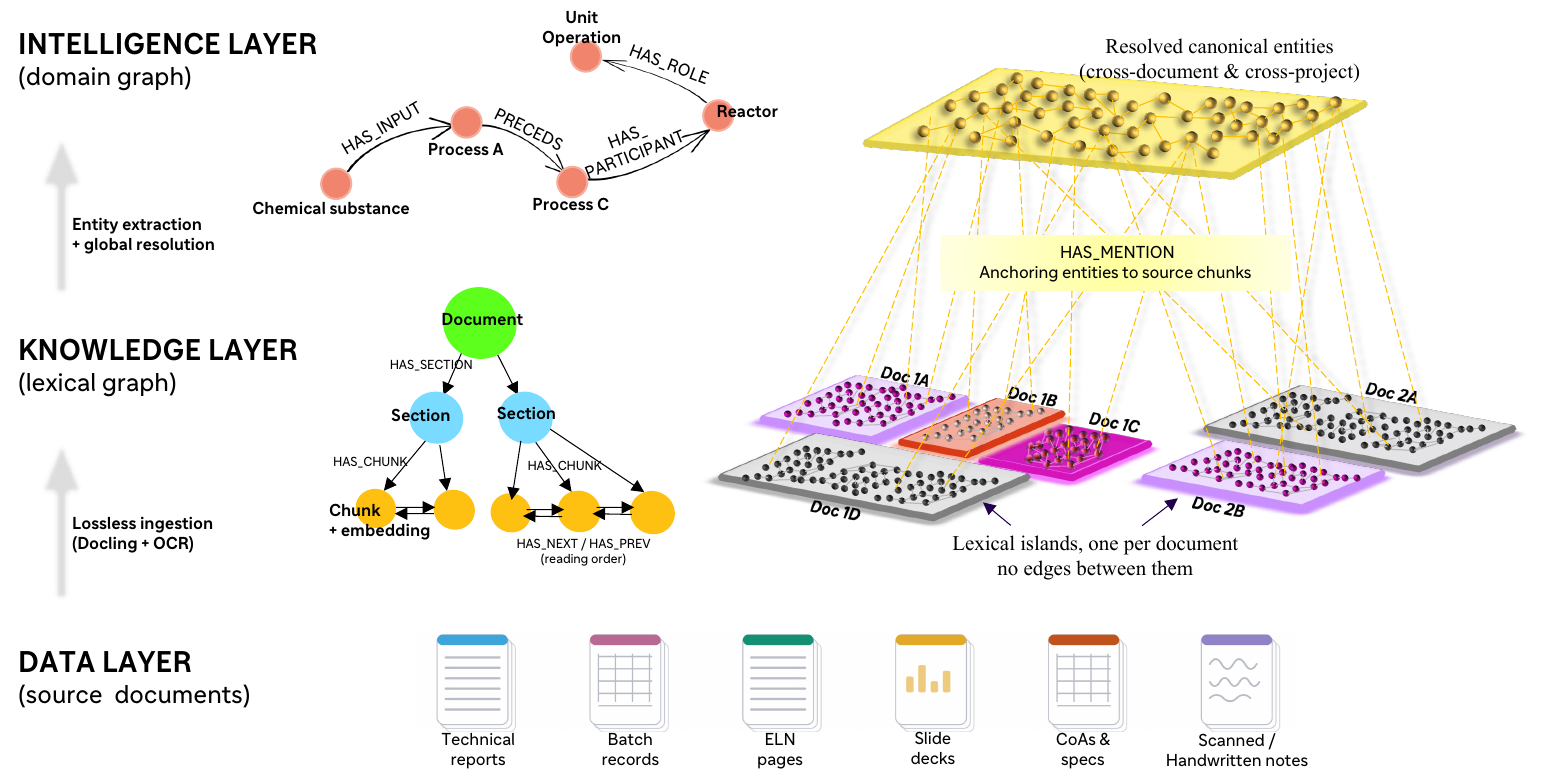}
  \caption{The multi-layer architecture: the data layer holds raw documents in heterogeneous formats, disconnected and without shared structure. Lossless ingestion of the documents produces the knowledge layer (lexical graph), in which every document becomes its own Document\,$\to$\,Section\,$\to$\,Chunk tree joined by HAS\_SECTION and HAS\_CHUNK edges, with HAS\_NEXT preserving reading; the knowledge layer is well suited to keyword and passage questions but carries no edges between documents. Entity extraction, based on a domain ontology, and global resolution produce the intelligence layer (domain graph), whose canonical entities are anchored to the chunks that mention them by HAS\_MENTION edges.  The intelligence layer merges lexical islands across documents and projects into one queryable graph.}
  \label{fig:figure2}
\end{figure}

A dual-layer modular architecture is designed to address a range of queries,
from keyword-oriented questions posed by an expert to process-design
decisions made by a project lead. \autoref{fig:figure2} summarizes the guiding
vision, in which an end-to-end pipeline converts raw documents into an
intelligence layer via an intermediate knowledge layer. At the bottom, a data
layer comprises raw, heterogeneous documents and data files from various sources, scattered across databases with no connection among them. Most biopharmaceutical organizations currently operate at this stage, where an LLM-based in-house assistant provides document summaries and answers predominantly keyword-oriented questions.

Lossless ingestion of these raw files produces an organized and machine-readable lexical graph (also called a knowledge layer), in which each file exists as an independent node, expandable to its sections and chunks (text, tables, and figures) with hierarchy preserved. This intermediate knowledge layer prevents knowledge loss while enabling AI-assisted question answering and retrieval.

The intelligence layer, built on top of the lexical graph, is an ontology-driven
graph. It contains the real-world concepts of a specific domain, in this case process chemistry, extracted from the chunks of the knowledge layer according to an applied class vocabulary. We refer to this level as the \emph{intelligence layer}, to the graph it holds as the \emph{domain} graph, and to the agent that operates on it as the \emph{domain} agent. The intelligence layer entities, while anchored to their source chunks, form a global graph that unifies domain-specific contents not only across documents within one project, but also across multiple projects. It is from the intelligence layer that patterns embedded in the connections between documents are revealed from a bird's-eye view, turning data into intelligence and ultimately decisions. The following sections describe the construction of each layer.

\subsection{Corpus and document heterogeneity}
\label{sec:corpus}

Although the platform has been tested internally on multiple CMC programs, each with 300+ technical documents, here we report quantitative results on a smaller testbed drawn from a discontinued Sanofi small-molecule program (the RIPK1 inhibitor oditrasertib; \cite{leszczak_process}). The testbed comprises 38 documents: 31 \verb|.docx| snapshots representing dated checkpoints of two living technical reports (21 checkpoints of one report and 10 of the other), one standalone \verb|.docx| appendix on solvent properties, and six \verb|.pdf| slide decks and appendices. Together, they contain more than 600 pages and roughly 52{,}000 words of technical content. The 31 snapshots cover 1-2 years of development efforts, with the two main reports growing from approximately 2{,}000 to 26{,}000 words (10--31 pages) and from 2{,}100 to 25{,}500 words (1--108 pages), respectively. The reports discuss process development, analytical measurement, and stability data in six sequential steps of chemical synthesis.

This emphasis on document lineages, rather than independent files, is by design. Version proliferation is one of the specific failure modes that this work targets: while a simple RAG system can support a user looking up keywords and specific details from an independent document, a process expert usually looks for information across a lineage of revisions, where conditions, batch identifiers, and conclusions shift between checkpoints, and the question that matters is often what changed and when. A corpus of independent documents would test a strictly easier problem than the one practitioners actually face. The testbed also contains genuinely independent material (six slide decks and appendices plus a standalone solvent-property document) covering both within-lineage and across-document cases.

Although the testbed discussed in this paper consists of documents in English, the platform is built to handle messy corpora, i.e., a mixture of digital and handwritten document scans in multiple languages (German, French, and Chinese), much of which is not machine-readable as received. This heterogeneity, typical of a technology transfer, is exactly what a practical ingestion pipeline must absorb.

\subsection{Knowledge layer: lexical graph}

Documents are parsed with Docling \citep{livathinos2025docling,
auer2024doclingreport}, which recovers layout, tables, and reading order, and
can normalize scanned content with optical character recognition (OCR) --- with optional translation to English for non-English source documents. Docling ingests PDF, DOCX, PPTX, XLSX, HTML, Markdown, AsciiDoc, CSV, and image inputs natively, so a single pipeline absorbs the full range of source formats with no pre-conversion step. Each document is written in a Neo4j property graph \citep{scifo2023graph} as a
Document\,--[HAS\_SECTION]\,$\to$\,Section\,--[HAS\_CHUNK]\,$\to$\,Chunk hierarchy, where sections nest to reflect the document's heading structure, and HAS\_NEXT edges link consecutive chunks within each section to preserve reading order (see the lexical graph in \autoref{fig:figure2}).

Chunks are element-aligned rather than produced by a fixed-width sliding window:
each Docling element becomes one chunk, so no overlap parameter is required and no element is split across chunk boundaries. Oversized tables are the exception and
are sub-chunked by row windows. Every chunk carries its text or table payload;
chunks with at least 25 characters of extractable text also carry a 1024-dimensional dense embedding (Amazon Titan Text Embeddings V2), stored on the node alongside the graph. Sub-threshold fragments and figure payloads are retained verbatim but not embedded, since they carry no retrievable prose. Retrieval combines the search for exact and full-text terms (Lucene BM25 over a full-text index of raw and
normalized text) with vector similarity (approximate nearest neighbor over a
cosine vector index). The result is a modular end-to-end pipeline that ingests an entire multilingual, mixed-modality corpus and makes it queryable. Within the lexical layer, each document is a self-contained Document\,$\to$\,Section\,$\to$\,Chunk tree. These trees form \emph{lexical islands}: excellent for locating information inside one document, but disconnected from one another.

\subsection{Intelligence layer: domain graph}
\label{sec:intelligence-layer}

The domain (ontological) graph is constructed from the same extraction JSON
files produced during lexical ingestion, so it does not add a new parsing burden. For each chunk, an LLM proposes ontological entities (each with a name, type,
description, aliases, and declared process inputs and outputs) typed against a
class vocabulary of chemical substance, planned process, unit operation,
equipment, and role. This vocabulary is aligned with OPC/PROCO and the
material/process distinction of the Basic Formal Ontology (BFO) \citep{he2020allotrope, allotrope2024afo, arp2015bfo}, and is expressed as a prompt-level schema together with a type-to-BFO-category map. We note explicitly that no OWL artifact is loaded and no description-logic reasoner or SPARQL engine is used here; the ontology functions here as a controlled vocabulary and a consistency constraint, rather than as an inference engine. 

The domain graph pipeline (i) generates entities (chemical\_substance, planned\_process, unit\_operation, equipment, role) and edges (PRECEDES, HAS\_SPECIFIED\_INPUT, HAS\_SPECIFIED\_OUTPUT, HAS\_ROLE, HAS\_PARTICIPANT, PART\_OF, IS\_A, HAS\_FUNCTION) and stores them as node--edge--node triples; (ii) performs global entity resolution to create a canonical entity registry, so that the many surface forms of a chemical substance collapse to one node (for instance: MeCN, ACN,  and acetonitrile); and (iii) assembles the domain graph, anchored back to the lexical layer through HAS\_MENTION edges linking each canonical entity to every chunk that mentions it. 

This anchoring keeps domain graph answers verifiable, such that an entity or relation reached by a HAS\_MENTION edge can be traced to the exact source passages supporting it, rather than only to a corpus-level summary of where such evidence might be. We refer to this design principle as \emph{provenance anchoring}, and note that not every entity in the registry includes a HAS\_MENTION anchoring edge. For instance, an LLM may assign a synthesized process label (e.g., \verb|S3_reaction|) that never appears verbatim in any chunk; the entity can also be auto-promoted from a cross-document input/output (I/O) reference and therefore never encounters raw text; or the entity is mentioned only under an abbreviation that was not recorded as an alias. Note that the entities without HAS\_MENTION edges are still reachable by traversing one relation-hop from a directly anchored neighbor in the domain graph, yielding 98.9\% effective provenance coverage in this case (98.9\% of entities are either directly anchored or one hop from a directly anchored entity).

This structure (direct anchor and one-hop reachable) enables corpus-wide answers  to be checked and helps the router agent move a question between the domain graph and lexical graph without losing the ability to verify either answer against text. For a regulated, auditable environment, such as GxP process development, dossier preparation, or deviation investigation, this property is consequential. In particular, a wrong answer that is traceable can be caught and corrected, while a wrong answer with no path back to source text cannot be distinguished from a right one without re-doing the underlying research. 

Entity resolution proceeds by union-find over document-local mentions, blocked by
entity type. Merges come from three sources in order of precedence: exact name and
alias matching; expansion through a curated process-chemistry synonym table
covering solvents, reagents, catalysts, and equipment; and, as a conservative
backstop, cosine similarity above 0.92 between entity vectors formed by mean-pooling the embeddings of the chunks in which each entity appears. Guards block chemically invalid merges. No LLM adjudication is used at this stage, so resolution is reproducible given fixed extraction output. A subsequent consistency pass checks that all members of a cluster agree on their BFO category, splitting the cluster where they do not. The surviving category is carried forward onto the canonical entity and written to the graph, so every entity is queryable by BFO category as well as by OPC/PROCO class.

\subsection{Reasoning: agentic workflow}
\label{sec:agents}

The reasoning incorporates three LLM agents. Two are implemented with the DSPy ReAct paradigm of interleaved reasoning and acting \citep{khattab2023dspy, yao2023react} and query a Neo4j graph database. The Lexical\_ReAct agent operates on the lexical graph and is assigned a budget of twelve reasoning iterations and eight retrieval tools, covering multiple retrieval dimensions of meaning, keywords, scope, and navigation within the Document\,$\to$\,Section\,$\to$\,Chunk graph, all issued as Cypher queries against Neo4j. Lexical\_ReAct excels at precise, single- or few-document questions, including exact keyword and full-text lookup. It is confined to chunk and document nodes with no path to canonical entities. 

The Domain\_ReAct agent traverses the canonical entity graph and excels at aggregation, exclusivity, overlap, and connectivity questions that span the corpus. It is equipped with a single tool, which reads the question and dispatches it to one of the parametrized Cypher templates over the edges that carry process-chemistry semantics. The agent selects among templates rather than freely authoring Cypher from scratch. Each template encodes the correct edge direction and parameter binding for its respective query type, which prevents the class of errors that arise when an LLM constructs graph traversals ad hoc. 

The third agent is a router, which selects between the Lexical\_ and Domain\_ReAct agents, according to the input question. It can also dispatch both agents, when a question needs resolved structure and passage detail together. Routing is deliberately two-stage and cheapest-first. The first stage involves a set of high-precision deterministic rules (LLM-free) to resolve the unambiguous cases. For instance, a question requesting a quantity or lot number is passage-level and goes to the lexical agent, whereas one asking what is unique to, shared between, or aggregated across documents goes to the domain agent. If none of the rules fires, the router makes a single tool-free LLM call to classify the question.  Because the domain graph is provenance-anchored, a domain graph answer can usually drop down to the
underlying chunks to ground or expand itself.

All agents are built with DSPy (\verb|dspy-ai| $\geq$ 2.5), run on AWS Bedrock with Anthropic Claude Sonnet 4.5 at a decoding temperature of 0.2 and a 2{,}048-token generation limit, and query a Neo4j graph database (Neo4j Desktop 2026.05, Python driver 5.x). 

\subsection{Benchmarking}

The benchmarking approach in this paper is designed to address the key gaps in reporting QA evaluation for RAG systems, and to introduce a deployment-fidelity benchmark that complements existing competence benchmarks. First, when the corpus involves proprietary programs, crucial questions must target the internal process history that no pretrained model could have memorized. This is inherently uncaptured in public-corpus benchmarks, where a model may answer a question correctly not just through improved passage retrieval, but also from its parametric memory (memorization confound). A recently reported benchmark for public settings is ChemRAG-Bench \citep{zhong2025chemrag}, where the RAG system outperforms direct inference by an average of 17.4\% on 1,932 questions assembled from four public datasets (larger models gained comparatively little improvement, consistent with memorization of those public sources). Here, we introduce a comprehensive benchmark tailored to settings with a proprietary corpus. Second, the benchmarking approach should account for all critical steps of an end-to-end knowledge graph pipeline, which includes graph construction, retrieval, and inference jointly. In this context, the information extraction from CMC documents is evaluated against annotated ground truth and reported in \citep{mann2023susie}. However, extraction quality does not guarantee reliability at the retrieval and inference stages. Our benchmarking approach converts an aggregate accuracy figure into a map of where the system can and cannot be trusted, thereby illuminating a failure taxonomy. Third, the quality and design of the question bank can significantly influence the reported QA score. We ensure that the questions probe corpus-specific knowledge that an SME would actually ask by generating questions from the corpus extraction JSON itself via an agentic AI workflow. The following subsections describe the workflow for question design and the three-tier assessment of the lexical graph layer.

The first step is to design questions that resemble those asked by an expert in the domain. We built a question-designer agent that reads the per-document extraction JSON and produces multiple-choice questions (MCQs) in a fixed schema: a stem, four options (one correct), the correct letter, an explanation, an evidence quote, the evidence location and source document, and a question-type label (procedure, results,
comprehension, process conditions, analytical method, and others). The questions are
generated in four cohorts according to anchoring mode and difficulty. An
\emph{anchored} question names a reference point (``In report~X, what was the value of \ldots{}?''), whereas an \emph{unanchored} question references no document or section (``What method is recommended for\ldots{}?''). \emph{Hard} questions are longer, must combine at least two evidence spans, or paraphrase rather than quote, and use distractor options that share vocabulary with the correct option. \emph{Comparative} questions are generated over document pairs (``How did process~X change from Doc-A to Doc-B?'') with both documents in context and both filenames mentioned in the stem. The four cohorts are: easy-mixed (a mixture of anchored and unanchored), hard-anchored, hard-unanchored, and comparative.

One quiz file is generated per document, with the question count scaled by document length as $\mathrm{round}(10 \times w / 8000)$ for a document of $w$ words, bound between 5 and 30. A mirror-overlap filter controls distractor difficulty. The quiz files from each document are concatenated, shuffled, and undergo cross-document deduplication (rejecting questions whose content-word overlap exceeded 0.65, plus exact evidence-quote duplicates). Twenty percent of the questions are randomly reviewed by an SME for correctness and quality before any evaluation is run; in the end, 100 questions are preserved for an initial pilot and 505 are selected for the final evaluation run. Refer to Appendix~A for examples of each cohort.

Following question design, a three-tier evaluation of increasing stringency is implemented. Tier 1 (T1, MCQ accuracy) compares the answering agent's selected letter (out of four options) against the gold answer. T1 is binary, cheap, and reproducible, making it suitable for rapid regression testing. In Tier 2 (T2), the same questions are asked without exposing the four options to the answering agent. Next, an LLM judge scores the answering agent's free-form justification on a 0--5 rubric for factual correctness, alongside the cosine similarity (embed\_sim) between the agent's answer and the reference explanation. T2 is more costly than T1, but eliminates the guessing floor of T1. Note that the judge model defaults to the same model and provider as the answering agent, Claude Sonnet 4.5 on Bedrock. The authors have not evaluated whether this shared lineage biases the judge toward the agent's own phrasing (a documented risk in LLM-as-judge setups).  The Tier-3 layer involves a human SME familiar with the corpus, who reviews items where the T1-pass (correct letter selected) and the T2-pass (judge\_score >= 4) disagree. For these items, the SME assigns a failure mode and may override a judge score. For instance, T2 exposes ``misleading positives'', in which the right letter is selected in T1 but the underlying reasoning is wrong or insufficient. Furthermore, we report two contingency tables, one for T1 × T2 with Cohen's Kappa, and one for judge\_score × embed\_sim. The Tier 3 review follows a stratified sampling design rather than exhaustively hand-coding every disagreement: every item where Tier 1 and Tier 2 disagree is reviewed in full; a stratified sample of items that fail both tiers is reviewed for attribution; and a thin spot-check of items that pass both tiers screens for a mis-keyed gold answer (\autoref{sec:lexical-results}). We report T1 accuracy and T2 pass rate with Wilson 95\% confidence intervals.

\section{Results and Discussion}

\subsection{Graph composition}
\label{sec:composition}

\begin{figure}[!htbp]
  \centering
  \includegraphics[width=0.86\linewidth]{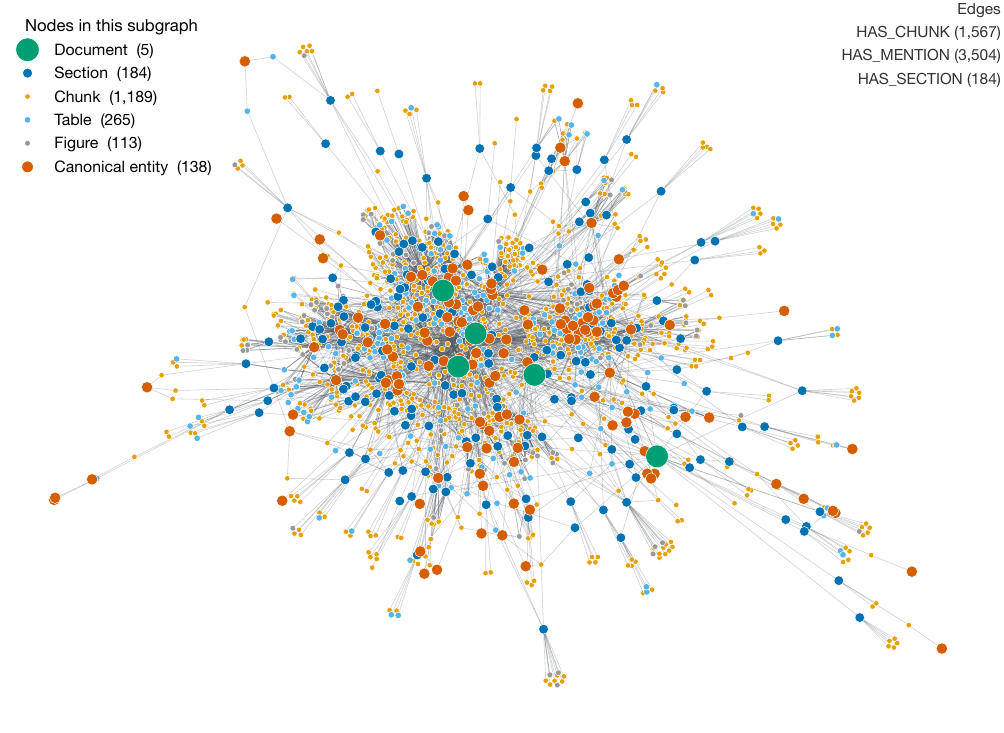}
  \caption{A portion of the dual-layer graph: the five documents whose chunk
  count is closest to the corpus median, with their sections and chunks and the
  138 canonical entities anchored to them. Document, Section and Chunk nodes
  (including the Table and Figure sub-labels) constitute the lexical layer;
  canonical entities from the domain graph attach to their source chunks by
  HAS\_MENTION edges, which is what bridges the per-document trees.}
  \label{fig:figure3}
\end{figure}

A dual-layer graph with 12{,}353 nodes and 45{,}471 edges is constructed from the complete processing of the testbed described in \autoref{sec:corpus}. \autoref{fig:figure3} shows a portion of the graph visualized in Neo4j.

The lexical graph comprises 38 Document nodes, 1{,}452 Section nodes joined by HAS\_SECTION edges, and 10{,}326 Chunk nodes joined by HAS\_CHUNK edges, with 9{,}178 HAS\_NEXT edges that preserve reading order. Of the chunk nodes, 1{,}925 are additionally labeled Table and 800 Figure, so that tabular and graphical payloads remain first-class retrievable objects rather than being flattened into prose; 93.2\% of chunks (9{,}620 of 10{,}326) carry a dense embedding, while the remainder consists of sub-threshold text fragments and figure payloads with no retrievable prose. 
The domain graph contributes 537 canonical entities: chemical substance (252),
planned process (153), unit operation (91), equipment (30) and role (11). The entities are anchored to the lexical layer through 22{,}688 HAS\_MENTION edges: 22{,}017 to individual chunks and 671 to whole sections. Of the 537 entities, 315 (59\%) carry at least one direct HAS\_MENTION edge to their respective chunks. The remaining 222 (41\%) entities, as described in  \autoref{sec:intelligence-layer},  are one relation-hop from a directly-anchored entity. These include: role-category entities (e.g. "catalyst", "solvent", "reagent", etc.) that are constructed as functional abstractions across many mentions; and per-step labels that the extraction agent inferred from surrounding context rather than from being named at one citable location. The relationships within the domain graph comprise HAS\_SPECIFIED\_INPUT (960), HAS\_ROLE (293), HAS\_SPECIFIED\_OUTPUT (264), PRECEDES (244), HAS\_PARTICIPANT (55), and PART\_OF (9), for 1{,}825 domain graph relations in total.

The lexical graph therefore contributes roughly 22 times as many nodes as the
domain graph, which is precisely the point of the dual-layer architecture. The
lexical graph preserves every piece of information for keyword matching, precise term lookup, and semantic search, while the domain graph distills the corpus into a navigable structure tailored to the domain expert (in this example, a process chemist). It is through this explicit, easily navigable domain graph that corpus-wide patterns are revealed. Note that the 22x ratio is a property of the entire corpus and is not reproduced in the \autoref{fig:figure3} subgraph. The ratio for this 5-document slice is smaller (138 canonical entities anchored to five document chunks, versus 1{,}756 lexical nodes in the same slice). Canonical entities are corpus-global, so all 537 are potentially reachable from any subset of documents, whereas the lexical nodes belonging to five documents are exclusive to those five documents. A small slice can therefore over-represent the domain layer, and the ratio approaches the corpus-wide
figure only as the document set approaches the full corpus. 

The global entity resolution on this corpus collapses 3{,}236 document-local mentions into 537 canonical entities. Of the total of 2{,}699 merges, 2{,}396 (88.8\%) are achieved through exact name, alias, or curated-synonym matching; 248 (9.2\%) through promotion of process-chemistry roles into entities with shared roles; and only 55 (2.0\%) through embedding similarity. This indicates a highly deterministic resolution, with the embedding step acting as a conservative backstop rather than the primary mechanism. Of the 537 canonical entities, 238 are singletons that appear in a single document and were never merged. No LLM adjudication is employed at this stage, which ensures that the resolution is reproducible given a fixed extraction output. A subsequent consistency pass in the pipeline checks that all members of a cluster agree on their BFO category, splitting the cluster where they do not; on this corpus, it examined 299 clusters (537 entities minus 238 singletons) and split none. The surviving category is carried forward onto the canonical entity and written to the graph, so that all 537 entities are queryable by BFO category as well as by process-chemistry class.

\subsection{Graph reproducibility}
\label{sec:reproducibility}

The construction of lexical graphs for the corpus studied here is deterministic, since Docling parsing, chunk creation, and Titan embedding do not involve an LLM call. However, entity extraction for domain graph construction can introduce nondeterminism. While comprehensive reproducibility testing is beyond the scope of this paper, the following observations are reported: repeating entity extraction from the chunks of one document, at the default temperature of 0.2, yielded 94 and 91 entities, respectively; repeating the experiment with the temperature pinned to zero and Python's hash randomization disabled yielded 83 and 86, respectively. We note that, in our analysis, the decoding temperature was the only available generation-time control parameter, and the sampling seed for the model was not exposed by Bedrock. Although two runs per condition are insufficient to isolate a true temperature effect from run-to-run noise, repeated runs within each condition still produced differing entity counts, descriptions, and chunk anchors, demonstrating that pinning temperature alone did not render extraction reproducible. We hypothesize that the residual variation reflects non-determinism in the inference service itself, which temperature does not control, and that this is plausibly amplified by the windowed multi-call structure of our extraction agent, where an early divergence could alter the near-duplicate consolidation that follows. Everything downstream of entity extraction, namely, global entity resolution, BFO validation, edge remapping, canonical-ontology construction, retrieval, and scoring, does not contain an LLM call and is reproducible given fixed extraction output. For regulated use, this is the gap that must be closed, requiring either a seeded inference endpoint or an extraction pass that is verified against the source text rather than simply trusted.

In the following section, the three-tier benchmarking approach is applied to the lexical graph and its ReAct agent; the limitations are then highlighted through a failure taxonomy. We subsequently discuss how the domain graph can complement the identified limitations.

\subsection{Lexical agent: reliability and limits}
\label{sec:lexical-results}

\begin{figure}[!htbp]
  \centering
  \includegraphics[width=\linewidth]{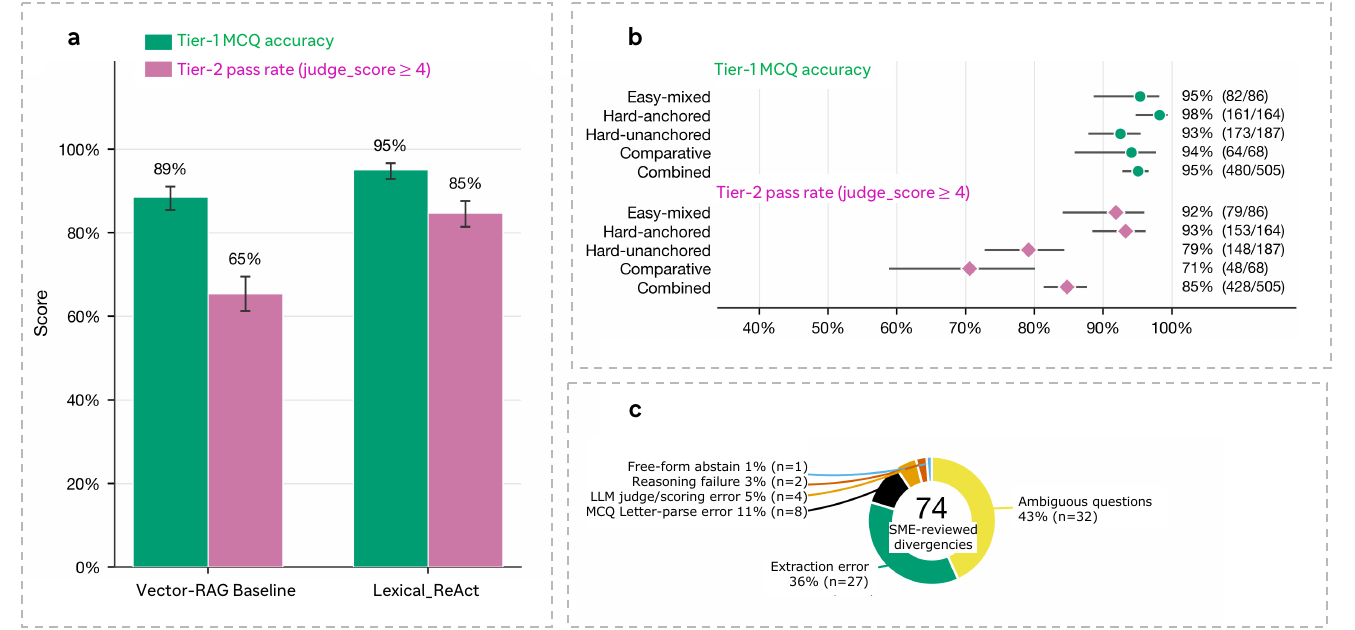}
  \caption{(a) Tier-1 multiple-choice accuracy (green) and Tier-2 pass rate (purple), defined as a correct T1 letter, and Tier-2 judge score $\geq$ 4, respectively, for the
  vector-RAG baseline and the Lexical\_ReAct based on 505-question
  curated bank. Error bars are Wilson 95\% confidence intervals.
  (b) T1 and T2 scores broken down by the four question cohorts. (c) Failure
  taxonomy over the 74 Tier-1/Tier-2 divergent items sent to Tier-3 SME review
  under a stratified sampling design (\autoref{tab:t1t2contingency})}
  \label{fig:figure4}
\end{figure}

A preliminary assessment of the Lexical\_ReAct based on 100 questions (designed in easy mode with mixed anchoring) revealed failure patterns in both retrieval and reasoning that guided agent prompt refinement. The final evaluation was run on a separate set of 505 questions that span the four cohorts of easy-mixed (86), hard-anchored (164), hard-unanchored (187), and comparative (68). The Lexical\_ReAct results from T1 MCQ accuracy and T2 pass rate (judge\_score >=4) are shown in (\autoref{fig:figure4}a) with green and purple bars, respectively. T1 accuracy is 95.0\% (480/505, Wilson 95\% CI 92.8--96.6) and T2 pass rate is 84.7\% (428/505, CI 81.4--87.6). The Lexical\_ReAct accuracy is assessed by comparing its scores with those from a baseline vector-RAG (88.5\% at T1 and 65.5\% on T2 pass rate) on the same 505-question bank. The Lexical\_ReAct outperforms the baseline RAG by 6.5\% on T1 and 19.2\% on T2. The improvement is explained by two mechanisms: first, the Lexical\_ReAct has access to the neighboring chunks via the lexical graph and a graph traversal tool to exploit it; second, the Lexical\_ReAct employs an iterative reasoning and acting loop as part of the ReAct paradigm, while the baseline RAG is loop-free. Note that the other variables, including the embedding model, chunk corpus, hybrid retrieval index, LLM prompt, answer parser, and LLM judge are kept constant.  

\autoref{fig:figure4}a also shows lower accuracy with the stricter Tier-2 assessment compared to Tier-1, in both the baseline RAG and Lexical\_ReAct. This highlights the \emph{recognition--generation gap} with the answering agent. In Tier-1, the agent can pick the correct option out of four, with a 25\% guess floor and elimination available even without full understanding (\emph{recognition}). However, in Tier-2, the agent must state the answer in prose against the gold explanation without a guess floor (\emph{generation}). The recognition--generation gap is notably smaller in the Lexical\_ReAct case compared to the baseline vector-RAG. These results emphasize how the final benchmarking scores are influenced by the type of QA approach employed (MCQ vs freeform judgment).

The other factors that affect the final scores are the question type and quality. An informative pattern is revealed by the cohort breakdown in \autoref{fig:figure4}b: T1 MCQ accuracy scores are high and relatively stable across all four cohorts. However,  T2 metric scores fall more significantly across the unanchored and comparative cohorts, which include the type of questions that require synthesizing evidence spread across multiple documents, i.e., a retrieval challenge for the answering agent. The score drop from T1 to T2 across the comparative cohort was even sharper for the baseline RAG (94\% to 19\%). \autoref{tab:t1t2contingency} lists the contingency of Tier-1 $\times$ Tier-2 for each cohort, illuminating the raw agreement (pass/pass and fail/fail cases) and the disagreement cases (T1-only and T2-only). Cohen's kappa is also quantified for each row to reflect the chance floor when both tiers mostly pass (0.17--0.32 reflecting low-to-moderate). Once the agreement expected from each tier's own marginal pass rate is subtracted out, Tier 1 and Tier 2 are shown to be measuring related but distinguishable capabilities rather than one being largely redundant with the other.

\begin{table}[!htbp]
  \centering
  \caption{Tier-1 $\times$ Tier-2 contingency, per cohort and pooled. T2 pass
  uses $\text{judge\_score} \geq 4$ rule (a materially complete and correct free-from answer); Cohen's kappa between T1 correctness and this stricter T2 pass}
  \label{tab:t1t2contingency}
  \footnotesize
  \setlength{\tabcolsep}{4.2pt}
  \begin{tabular}{lrrrrrrrr}
    \toprule
    \textbf{Cohort} & \textbf{$n$} & \textbf{T1} & \textbf{T2} &
    \textbf{pass/pass} & \textbf{T1-only} & \textbf{T2-only} &
    \textbf{fail/fail} & \textbf{$\kappa$} \\
    \midrule
    Easy-mixed        & 86  & 95.3\% & 91.9\% & 77  & 5  & 2  & 2  & 0.32 \\
    Hard-anchored      & 164& 98.2\% & 93.3\% & 152& 9  & 1  & 2  & 0.26 \\
    Hard-unanchored    & 187 & 92.5\% & 79.1\% & 141 & 32 & 7  & 7  & 0.17 \\
    Comparative        & 68  & 94.1\% & 70.6\% & 47  & 17 & 1  & 3  & 0.17 \\
    \midrule
    Combined           & 505& 95.0\% & 84.7\% & 417& 63 & 11 & 14 & 0.22 \\
  \bottomrule
  \end{tabular}
\end{table}

It is worth noting that Tier-2 score quality depends not just on the performance of the Lexical\_ReAct, but also on the accuracy of the LLM judge. To ensure judge accuracy, the embedding similarity (cosine similarity between the free-form answer and the gold explanation) is also captured as part of the assessment. We note that the LLM judge\_score and embedding similarity are imperfect proxies for one another; this is shown by cross-tabulating judge score against embedding similarity, split at each cohort's median similarity (Appendix A). The paraphrase trap quadrant—high embedding similarity but low judge score—captures correct-sounding answers that are factually wrong, a failure mode that similarity metrics alone cannot detect. Across all 505 scored answers, 22 instances fall in this quadrant, concentrated on hard-unanchored and comparative questions where reasoning must synthesize evidence across documents. Judge score and embedding similarity correlate only moderately (Spearman $\rho$=0.30, p<0.001), underscoring that surface similarity to the gold standard is not a substitute for semantic correctness. A full contingency table is provided in the appendix.

The disagreement cases between the first two tiers comprise 74 of the 505 questions (63 T1-only and 11 T2-only) . These questions plus a 30\%-stratified sample of items that fail both tiers are reviewed by an SME in Tier-3. The failure taxonomy is revealed in \autoref{fig:figure4}c.  \emph{Question-quality issues} are the largest category (43\%, n=32). These items were judged by the SME as either ambiguous, internally inconsistent, or resting on a table or entry number that disagrees with its own gold explanation. These were followed by \emph{extraction errors} (36\%, n=27), \emph{letter-parsing errors} (11\%, n=8), \emph{judge scoring errors} (5\%, n=4), \emph{reasoning failures} (3\%, n=2), and a single residual \emph{free-form abstention} (1\%, n=1). Most of the disagreement between the two tiers is attributable to the questions themselves, not to the agent's retrieval or reasoning. This limitation can be addressed by further improving the question-designer LLM agent. Letter-parsing errors refer to cases where the freeform answer in Tier-2 was accurate, while a wrong letter was picked in Tier-1. We believe this is caused by excessive similarity between two options and could be reduced by refining the questions (an artifact of the MCQ format rather than of retrieval). The dominance of extraction errors, compared to reasoning failures, is also an encouraging sign for future improvements to the platform. A closer look reveals that the extraction errors are concentrated on hard-unanchored questions (45\% of the reviewed divergences) for which the answering agent has few clues to find the evidence location(s). This can become significant with a corpus size much larger than the one studied here. 

From a practical perspective, comparing Tier-1 and Tier-2 reveals two important considerations for reporting QA-based benchmarks: first, depending on the question type and quality, reporting only Tier-1 MCQ accuracy can overestimate a pipeline's reliability. This might not be an issue when the benchmarking is primarily based on easy questions (such as keyword-oriented ones or those hinting at a specific document/section/chunk); yet, assessments based on longer questions, with their evidence scattered across multiple locations, can result in divergence between Tier-1 and Tier-2. The Tier-1 approach is the better metric for comparing platforms head-to-head, because it is reproducible and cheap; whereas Tier-2 can reveal the recognition--generation gap and is a better instrument for locating failure mechanisms within one platform. We emphasize the importance of Tier-2 especially for regulatory applications such as those encountered in the pharmaceutical industry.

\subsection{Domain agent as the complement}
\label{sec:domain-agent}

Systematic benchmarking of the knowledge layer (lexical) in \autoref{sec:lexical-results} illuminates the question patterns where the answering agent is likely to fail. In fact, the failure is expected to grow when the comparison spans not a pair but many documents, or an entire project corpus of 300+ documents. In the context of CMC synthetics, a process expert is often interested in the evolution of a substance, process, or impurity across a multi-step synthetic route or in the
commonality of specific materials or methods across steps or even projects. Answering such queries---even with multi-hop traversal---is extremely challenging for the lexical ReAct agent. As reported in \autoref{sec:composition}, the lexical layer for a corpus of 38 documents holds 11{,}816 nodes. We report that graph size can grow to $\sim$80{,}000 nodes when a corpus with $\sim$300 of internal documents is ingested using the same pipeline. One can therefore anticipate that relying only on lexical graphs, which have high precision but low domain specificity, would limit the use case of GraphRAG platforms for cross-project applications such as anomaly detection, or route designs based on historical insights. This limitation is addressed by anchoring a refined domain graph on top of the lexical graphs (see \autoref{fig:figure2}).     

\begin{figure}[!htbp]
  \centering
  \includegraphics[width=\linewidth]{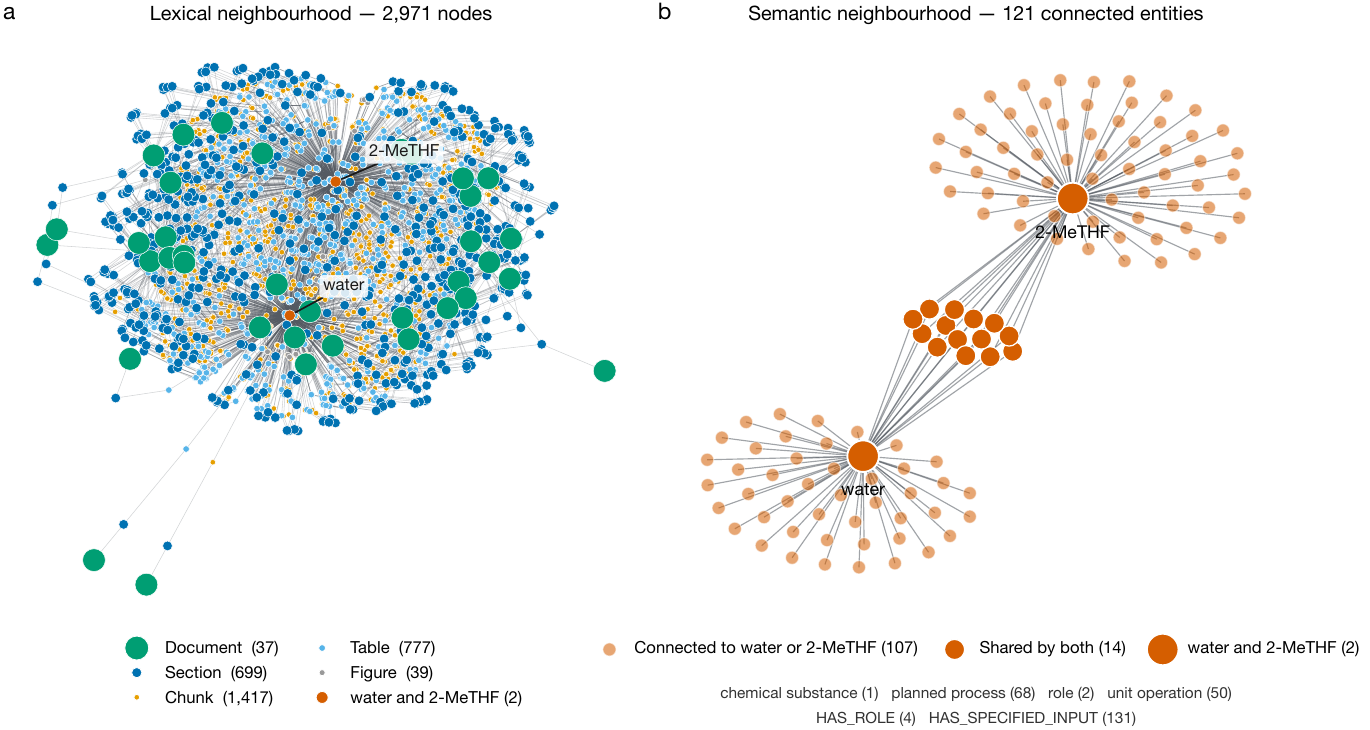}
  \caption{(a) lexical graph nodes connected to the water and 2-MeTHF canonical
  entities: 2{,}971 nodes spanning 2{,}233 chunks (of which 777 are tables) plus
  their parent sections and documents. (b) The 121 canonical entities connected to
  water or 2-MeTHF, with the 14 entities adjacent to both highlighted; these are
  the candidates from which the answer set is drawn. The compactness and contextual-awareness of the domain graph enables answering corpus-wide questions, such as ``what process steps have water and 2-MeTHF in common?'' in a single query. Node counts in the legends are computed, not transcribed.}
  \label{fig:figure5}
\end{figure}

Consider the  question ``Which process steps have water and 2-MeTHF in common?'' asked from Lexical\_ and Domain\_ReAct agents separately. \autoref{fig:figure5} contrasts (a) the lexical and (b) the domain neighborhoods around the water and 2-MeTHF entities. In the lexical graph, both entities are extremely high-degree concepts: together they are mentioned in 2{,}233 distinct chunks, which with their parent sections and documents form a neighborhood of 2{,}971 nodes. The lexical agent must review those passages a few at a time to assemble a list of process steps, and return an incomplete list of two items, since it cannot observe every co-occurrence at once. The domain agent instead has access to a clean domain graph in which 121 canonical entities connect to water or 2-MeTHF. Its single tool returns each
solvent's typed neighbors in one query apiece, rather than the passage-level
search the lexical agent must repeat; of the returned neighbors, 14 are adjacent to both solvents, with 5 of them as planned processes. In this case, the domain agent retrieves a clean list of 5 processes while the lexical agent answers an incomplete list.
The domain agent also returns each entity with its source chunks, via the HAS\_MENTION anchoring edges, so the answer remains checkable against the documents. The same traversal generalizes directly to anomaly and comparison questions---such as``which batch differs from the rest" and ``what changed between versions''---which are impractical for chunk-by-chunk retrieval but natural over a resolved entity graph.

To this point, we have described how an agentic platform with two ReAct agents can
address different query types, from keyword-oriented questions to corpus-wide aggregation under a certain ontology. At a higher level, we implement a router agent that selects the retrieval path and the corresponding ReAct agent, or their combination, according to query type. We further designed a small set of  "unanchored/comparative" questions to compare the two agents head-to-head, scored by an SME. \autoref{tab:headtohead} lists a representative subset of questions and the selected agent with the most accurate answer. The recurring success/failure modes are informative for a pattern to define for the router agent. In particular, the domain agent does better on most cross-document items; whereas the lexical agent is more accurate on judgment-and-narrative items. On corpus-wide and cross-document questions, the lexical agent typically returns a correct but incomplete list. One example question is "which process steps have acetonitrile as an input material?"

\begin{figure}[!htbp]
  \centering
  \includegraphics[width=0.75\linewidth]{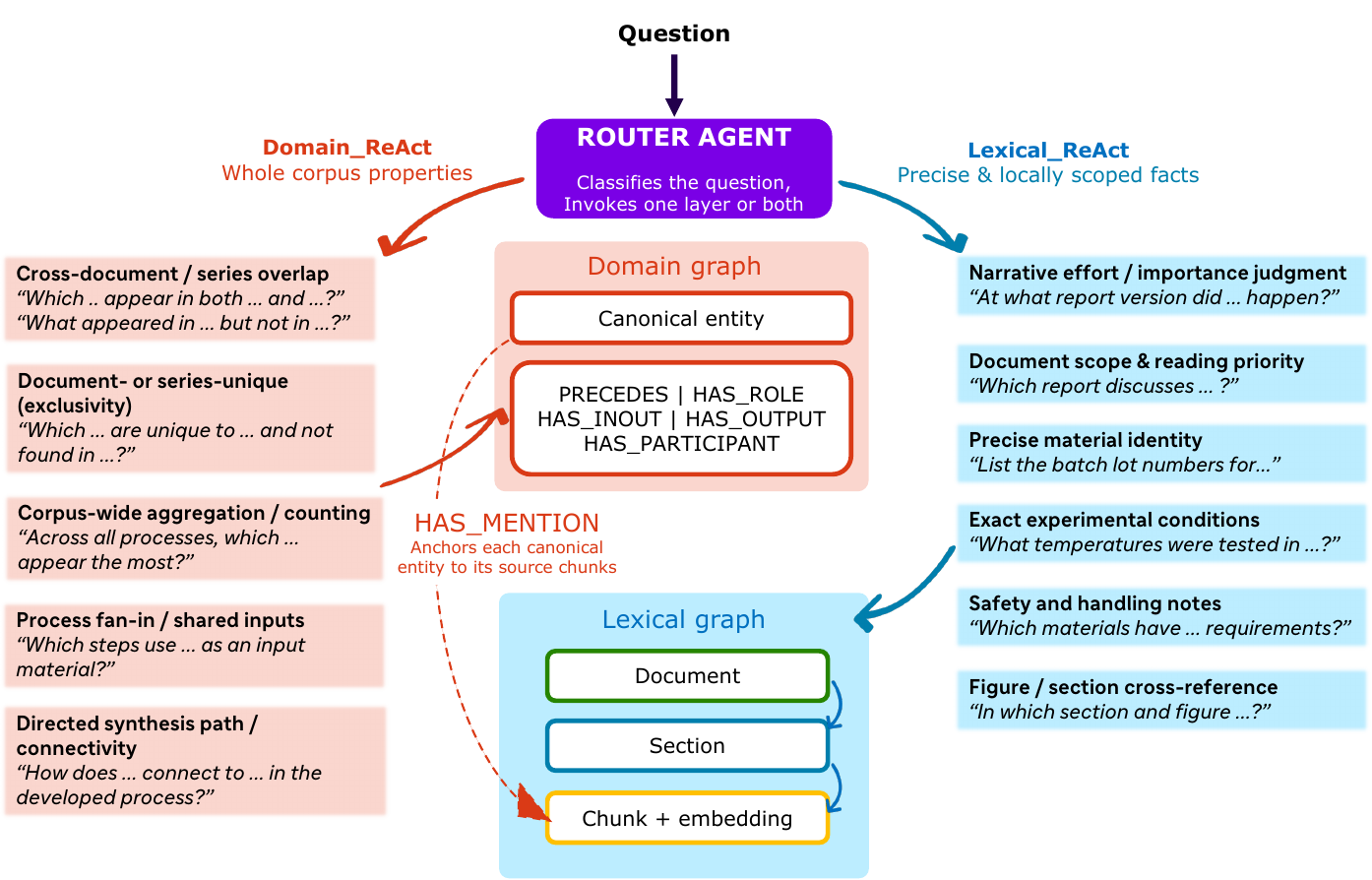}
  \caption{Question routing and the questions each layer serves. Every question
  enters the router agent, which invokes the domain agent (left), the lexical
  agent (right), or a hybrid mode holding both toolsets. The domain agent
  operates on process-chemistry concepts (canonical entities) and answers
  properties of the corpus as a whole; the lexical agent retrieves precise, locally
  scoped detail from chunks and their embeddings. HAS\_MENTION runs from each
  canonical entity to the chunks that mention it, so a domain graph answer can
  usually be traced down to the passages that support it.}
  \label{fig:figure6}
\end{figure}

\autoref{fig:figure6} summarizes the sample CMC questions best suited to each
agent. The lexical agent is the right tool for precise, locally scoped needs: exact experimental conditions, material identity and batch lot numbers, safety and handling notes, figure and section cross-references, and narrative or content judgments. The domain agent is the right tool for questions whose answer is a
property of the whole corpus in the context of process chemistry: corpus-wide
aggregation and counting, document- or series-level uniqueness (exclusivity),
entity evolution across document versions, cross-document overlap, directed
synthesis-path connectivity, and process fan-in through shared inputs. The two
are complementary and a router agent can combine them.

\begin{table}[!htbp]
  \centering
  \caption{Representative questions comparing the domain and lexical agents
  head-to-head, from a small set of questions scored by a subject-matter expert.
  Cross-document aggregation, exclusivity, overlap and connectivity favor the
  domain agent; scope and narrative judgments favor the lexical agent.}
  \label{tab:headtohead}
  \scriptsize
  \setlength{\tabcolsep}{3pt}
  \renewcommand{\arraystretch}{0.95}
  \begin{tabular}{p{0.50\linewidth} p{0.09\linewidth} p{0.34\linewidth}}
    \toprule
    \textbf{Representative question} & \textbf{Selected agent} & \textbf{Basis} \\
    \midrule
    What process steps have water and 2-MeTHF in common?
      & Domain & Lexical list correct but incomplete \\
    \midrule
    Which process steps have acetonitrile as input material?
      & Domain & Lexical correct but incomplete \\
    \midrule
    Of all solvents across the documents, which top three were tested most?
      & Domain & Lexical correct but incomplete\\
    \midrule
    Which processes or chemicals are unique to report \#18?
      & Domain & Structural exclusivity query \\
    \midrule
    How does turbo-Grignard connect to compound~6 in the chemistry?
      & Domain & Multiple valid paths returned \\
    \midrule
    Which substances appear in both report series A and B?
      & Domain & Lexical partial; domain complete \\
    \midrule
    At what report version did the largest cyanation development occur?
      & Lexical & Passage reading \& reasoning required \\
    \midrule
    I can read only 3 of 38 reports---which ones do you recommend?& Lexical & Judgement about scope and narrative \\
    \midrule
    Which report best describes the iodometric titration?
      & Lexical & Domain abstained; lexical listed documents \\
    \midrule
    Which process uses dimethylacetamide and runs in a CSTR?
      & Tie & Both correct and complementary \\
    \midrule
    What are all chemical inputs to the SNAr cyclization step?
      & Tie & Both correct; lexical used a detailed table \\
    \bottomrule
  \end{tabular}
\end{table}

\section{Summary, Limitations and Outlook}
\label{sec:outlook}

We presented an agentic platform with a dual-layer knowledge graph that turned a
heterogeneous corpus of process-development documents into a queryable
institutional memory. A knowledge layer, based on a lexical or precision graph,
was built on lossless ingestion of a variety of document formats---digital,
scanned, handwritten, and multilingual---to enable both search by keyword and
search by meaning through hybrid vector-and-graph retrieval.

A three-tier benchmarking protocol was introduced for comprehensive evaluation of the knowledge (lexical)
layer. The assessment was based on 505 questions, designed by an LLM agent and validated by a human SME, over 38 development reports. We highlight the strengths and limitations of each assessment tier and emphasize the quality of the question bank for reporting benchmarking scores. The Lexical layer outperformed the vector-RAG baseline in both Tier-1 MCQ accuracy (95\% vs 89\%) and more significantly in Tier-2 pass rate (85\% vs 65\%). Tier-1 and Tier-2 contingency per question cohort revealed the failure taxonomy. While the lexical agent performed well on precise, locally scoped questions, it fell comparatively short on corpus-wide queries and unanchored questions, particularly those centered on domain knowledge.

Those limitations were partly addressed by an ontology-guided intelligence layer, a unifying
domain graph that was provenance-anchored to the lexical foundation. The domain graph benefits
from global entity resolution---a step essential in CMC process development---and
bridges otherwise isolated document chunks to answer the cross-document aggregation, comparison, and anomaly questions on which passage retrieval alone falters.

The architecture provides modularity and admits extension to other scientific areas. The lexical knowledge layer (Document, Section, and Chunk) is independent of project content, whereas the domain graph can adopt tailored ontologies
(small molecule, large molecule, vaccines, etc.) to serve domain-specific queries. We anticipate extending the current architecture from dual- to multi-layer by adding material and equipment databases together with literature and compound
data, in both internal and external formats.

We emphasize that a general-purpose AI chatbot and a knowledge graph are not interchangeable. While internal chatbots can summarize a handful of uploaded documents quickly, they cannot
hold an entire program in a bird's-eye view. In the context of CMC synthetics,  anomaly detection and
corpus-wide aggregation are arguably the highest-value CMC questions during scale-up and filing. Answering them requires the unified, resolved, cross-document structure that the domain graph provides. 

Several limitations bound the present study, and we group them by what would be
required to address each in future studies.

\emph{Evidence for the domain graph.} The three-tier protocol was applied to
the Lexical\_ReAct agent only. The domain graph was evaluated based on a small set of questions and was compared 
head-to-head with the Lexical\_ReAct agent to reveal retrieval patterns for the router agent. A domain-agent benchmark at comparable scale, and across more documents and multiple projects, is the highest-priority next
step. For regulated CMC applications, auditability matters as much as accuracy,
and benchmarking must be part of the system. Our protocol converted a reassuring headline accuracy into an actionable map of where the platform can and cannot be trusted, and the failure taxonomy explained why.

\emph{Isolating the graph's contribution.} The improvement of Lexical\_ReAct over the vector-RAG baseline is attributed to
the combination of the lexical graph and the agentic retrieval loop (\autoref{sec:lexical-results}),
not to graph structure alone, because the agent and the baseline also differ
in tool access, iteration budget, and prompt effort. An isolating experiment
--- using the same agent, prompt, and iteration budget, while toggling only
graph-structural navigation tools against flat retrieval --- must be run to distinguish the graph retrieval tool from the ReAct retrieval loop.

\emph{Reproducibility.} Extraction is not reproducible, and pinning the decoding
temperature to zero did not make it so: two extractions of the same document at
temperature zero, with hash randomization disabled, differed in entity count (83
versus 86) and in entity descriptions and chunk anchoring. The deterministic
downstream phases did reproduce exactly, given fixed extraction output. Closing the remaining gap needs something stronger than a temperature setting --- a seeded
inference endpoint, or a verification pass that checks each proposed entity against the source span rather than simply accepting it. This is a prerequisite for future regulated use.

\emph{Ontological depth.} The class vocabulary was applied as a prompt-level
schema and a consistency constraint, rather than as a loaded OWL artifact with a
reasoner. Only five classes were instantiated in practice. Properly loading the OPC/PROCO ontologies \citep{he2020allotrope} and running a description-logic reasoner would permit inferred relationships and formal consistency checking, neither of which we currently exploit. Richer ontology coverage---particularly around impurities, fate, and purge, and control strategy---would deepen the classes of anomaly that the graph can surface.

\emph{Reaction schemes.} Several benchmark questions had to be excluded because
their answers resided in figures that the pipeline does not read. A high-accuracy
vision-language model that converts reaction-scheme images into structured
representations (CDXML or CML) would close this gap; reaction schemes are dense
with mechanistic meaning and would improve domain graph fidelity during ingestion.

\emph{Entity-resolution precision.} The domain graph extraction inherits LLM error modes. An explicit grounding and verification pass in the style of ODKE+ \citep{khorshidi2025odke} and confidence-weighted canonicalization would tighten
domain graph precision. We detected two specific extraction quality issues that required correction prior to building the domain graph: under-populated solvent roles and over-tagging of strong bases as catalysts.

\emph{Router quality.} The rules for the routing agent were defined based on a head-to-head comparison of the Lexical\_ and Domain\_ReAct agents with two caveats worth mentioning: first, the question set was small and limited to one project's documents; second, the human SME selected the winner answer without blinding to which agent produced which answer. A future experiment should run the router over the 505-question bank, while recording its decision to select the answering agent alongside every answer. By doing so, a correctly routed question answered badly is charged
to the answering agent and not to the router. That work should follow the domain graph benchmark, since routing accuracy is only interpretable once both destinations are independently characterized.

\emph{Large Context LLM vs GraphRAG.} A modern long-context LLM can be an alternative approach to the GraphRAG pipeline discussed in this paper if the corpora size is small to moderate. Indeed, for a testbed of 52{,}000 (38 documents), a long-context LLM given the full corpus might score comparably to either the vector-RAG or Lexical\_ReAct. While the comparison is not tested here, we argue that the graph's motivating regime is found in multi-hundred-document programs mentioned in \autoref{sec:corpus}. A typical project of 300+ internal documents imposes challenges with cost, scale, structural reasoning, and updates when processed with flat retrieval.

Finally, the platform architecture and benchmarking protocol introduced here are not specific to CMC process development: the failure taxonomy is a general instrument for classifying why an MCQ-plus-judge benchmark's two tiers disagree, and the same failure codes apply unchanged to any RAG or agentic system scored the same way. More broadly, we expect biopharmaceutical organizations to move from siloed, document-centric knowledge management toward graph-based institutional memory, and we anticipate that shared benchmarking practices and process-chemistry ontologies will become necessary infrastructure for making such systems trustworthy, auditable, and comparable across the industry. Turning fragmented technical records into a queryable, provenance-anchored graph is, in this sense, the more direct route from document silos to process intelligence.

\section*{Author Contributions}

Reza Amirmoshiri conceived and led the platform, developed the lexical and semantic pipelines and the benchmarking, and wrote the manuscript. Faryad Sahneh contributed to the generative-AI methodology, platform architecture, and programming. Yasser Jangjou initiated the project roadmap, provided CMC process-development guidance, and use-case definition. All authors reviewed and approved the final manuscript.

\section*{Acknowledgments}

The authors thank Sebastien Paltrie \& Nicolas Rouyer from Neo4J -- for production grade graph rag solution development during an MVP for a commercial project. The authors thank Mohan Boggara and Robert Ronnback for their comments on the draft of the article. We thank Sanofi US for granting permission to publish this work.

\section*{Data Availability and Ethics}

The reported testbed belongs to a discontinued Sanofi small-molecule program (the RIPK1 inhibitor oditrasertib). The underlying documents contain proprietary information and cannot be released; the process patent for this
compound is available separately \citep{leszczak_process}. The larger internal corpus referenced for scale is confidential. No human-subjects or animal data was used.

\section*{Conflicts of Interest}

The authors are employees of Sanofi US. The authors declare no other competing interests.

\appendix

\section{Question generation}
\label{app:questions}

\subsection{Question-designer agent}

The question-designer agent reads the per-document extraction JSON and emits multiple-choice items in a fixed schema. The prompt is assembled per document from a common preamble plus two switchable rule blocks --- one for
anchoring mode and difficulty --- to generate four cohorts. The schema fields are: stem, four options, the
correct letter, an explanation, a verbatim evidence quote, the evidence location,
the source document, a question-type label, and a confidence score.

\noindent Two further mechanisms control item quality. A diversity hint injects the
previous five stems and the facets already covered for that document, instructing
the model to prioritize uncovered facets. A mirror-overlap filter rejects a
distractor only when its word overlap with the gold option is at least 0.78
\emph{and} its numeric tokens are identical \emph{and} its character-level
similarity ratio is at least 0.75; the numeric condition is what preserves
legitimately discriminating pairs such as ``3.5~min at 30~\textdegree C'' versus
``2.0~min at 50~\textdegree C''.

The following four items are taken unmodified from the evaluated question bank.
The correct option is marked ($\star$).

\paragraph{Easy-unanchored (question type: analytical method).}
\emph{What melting point was determined for Compound F by DSC analysis?}
\begin{enumerate}[label=\Alph*., itemsep=0pt, topsep=2pt]
  \item 75.0 °C
  \item 71.1 °C ($\star$)
  \item 65.8 °C 
  \item 69.4 °C
\end{enumerate}
\textbf{Evidence:} "The melting point for Compound F was found to be 71.1 °C, with an onset melting temperature of 69.4 °C." (Table 50 Discussion, \emph{TR-1-083-\#10})

\paragraph{Hard-anchored (question type: procedure).}
\emph{In TR-1-083-\#13, Table 51 details the preparation of the TFA salt of RA-X35. What solvent and volume were used to dissolve 1.0~g of RA-X35 before TFA addition?}
\begin{enumerate}[label=\Alph*., itemsep=0pt, topsep=2pt]
  \item 10~mL of 2-MeTHF
  \item 10~mL of acetonitrile
  \item 10~mL of MTBE ($\star$)
  \item 5~mL of heptane
\end{enumerate}
\textbf{Evidence:} ``RA-X35 input: 1.0~g; MTBE amount: 10~mL''(\emph{TR-1-083-\#13}, Table 51). (Note the stem names the document and the table, and every distractor is a solvent that appears elsewhere in the same report.)

\paragraph{Hard-unanchored (question type: process conditions).}
\emph{During the continuous quench implementation for the S'1--S2 demonstration
run, what pH value was maintained in the CSTR while delivering the aqueous acid
solution?}
\begin{enumerate}[label=\Alph*., itemsep=0pt, topsep=2pt]
  \item pH fluctuated between 11 and 13 during the run
  \item pH dropped to 5 after the first 30 minutes of operation
  \item pH stabilized at 7 after initial equilibration
  \item pH remained around 9 throughout the continuous quench ($\star$)
\end{enumerate}
\textbf{Evidence:} ``The reactor was stirred vigorously ($>$500~rpm) and a pH probe
was used to monitor the pH of the mixture, which remained around 9.'' (Section
4.3.1, K9-13 demonstration run). The stem names no document, so the agent must
locate the run by its technical description alone.

\paragraph{Comparative (question type: comparative).}
\emph{What major shift in research focus occurred between report series TR-1-2018 and
TR-1-2020?}
\begin{enumerate}[label=\Alph*., itemsep=0pt, topsep=2pt]
  \item All steps received equal attention in both reports
  \item Emphasis moved from Steps 1, 1', 2, and 5 to Steps 3 and 4 ($\star$)
  \item Focus shifted entirely to analytical method development
  \item Emphasis moved from Steps 3 and 4 to Steps 1, 1', 2, and 5
\end{enumerate}
\textbf{Evidence:} ``In the previous work package, TR-1-2018, emphasis was placed on
investigating S1, S'1, S2, and S5, while S3 and S4 were only preliminarily
investigated. In the work package TR-1-2020, we focus on \ldots{}''
(TR-1-2018 $\rightarrow$ TR-1-2020, Report Summary and Project
Background). Option D is the same claim with the direction reversed, which is the
distractor pattern the comparative generator is designed to produce.

\section{T1/T2 concordance and the paraphrase trap}
\label{sec:results-contingency}

\begin{table}[!htbp]
  \centering
  \caption{Judge score $\times$ embedding-similarity contingency, per cohort and
  pooled, at $\text{judge\_score} \geq 4$ threshold and each cohort's
  own median similarity split. Columns: \textbf{on-topic} = judge high, sim.\
  high (expected correct); \textbf{paraphrased} = judge high, sim.\ low
  (correct but divergent phrasing); \textbf{trap} = judge low, sim.\ high (the
  paraphrase trap: judge says wrong, embedding says similar); \textbf{off-topic}
  = judge low, sim.\ low (expected fail).}
  \label{tab:paraphrasetrap}
  \footnotesize
  \setlength{\tabcolsep}{4.2pt}
  \begin{tabular}{lrrrrrrr}
    \toprule
    \textbf{Cohort} & \textbf{$n$} & \textbf{med.\ sim.} &
    \textbf{on-topic} & \textbf{paraphrased} &
    \textbf{trap} & \textbf{off-topic} & \textbf{$\rho$} \\
    \midrule
    Easy-mixed        & 86  & 0.764 & 43 & 36& 0  & 7& 0.27 \\
    Hard-anchored      & 164& 0.765 & 78& 75& 4& 7& 0.17\\
    Hard-unanchored    & 187 & 0.713 & 83& 65& 11& 28& 0.28 \\
    Comparative        & 68  & 0.710 & 28& 20& 7& 13& 0.39 \\
    \midrule
    Combined           & 505& ---   & 232& 196& 22& 55& 0.28\\
  \bottomrule
  \end{tabular}
\end{table}

\section{Worked head-to-head example: shared solvents}
\label{app:transcripts}

This appendix gives the answers behind the example in \autoref{sec:domain-agent}, for
the question \emph{``Which process steps have water and 2-MeTHF in common?''}. Answers are reproduced as
returned, truncated for length.

\subsection{Lexical agent}

\begin{quote}\small
Two process steps have water and MeTHF in common:

\textbf{1. S2 --- Oxime Elimination to Nitrile (Reaction Step).} The elimination of
oxime 9 to form nitrile 7 occurs in MeTHF at 60~\textdegree C with 1.2~equiv.\ of
CDI. Water sensitivity was specifically assessed for this step to determine if it
could be telescoped from the oxime formation without removing water. Experiments
were conducted in MeTHF with water content ranging from anhydrous
($<$100~ppm) to saturated ($\sim$4\%). \emph{(Source: TR-1-2018\#8.docx,
section 3.2.4.2)}

\textbf{2. S3 --- SNAr Workup (Extraction Step).} The workup procedure for the SNAr
reaction involves DI water extraction of the crude reaction mixture in MeTHF.
Table 10 shows workup conditions for experiment SB01-71 using DI water extraction
at a 1:1 volume ratio to crude solution. \emph{(Source: TR-1-2018-\#2, Table 10)}
\end{quote}

\noindent The two steps returned are correct and each is properly cited. The answer
is nonetheless incomplete: the agent stops after two, because water and 2-MeTHF are
together mentioned in 2{,}233 chunks and it can only inspect a few at a time.

\subsection{Domain agent}

\begin{quote}\small
The following process steps have both water and 2-MeTHF (2-methyltetrahydrofuran) in common:
\begin{enumerate}[itemsep=1pt, topsep=3pt]
  \item \textbf{dgf03\_037\_purification\_3\_5\_dibromo} --- a purification unit
  operation that uses water and 2-methyltetrahydrofuran to purify 3,5-dibromo-4-chloropyridine
  \item \textbf{azeotropic\_drying\_methf} --- an azeotropic drying operation that
  uses water and 2-methyltetrahydrofuran along with magnesium sulfate to remove compound F
  \item \textbf{phase\_separation} --- a general phase separation operation with
  inputs including water, 2-methyltetrahydrofuran and various compounds
  \item \textbf{aqueous\_quench} --- a quenching operation that uses
  2-methyltetrahydrofuran as solvent with aqueous workup components
  \item \textbf{heptane\_backwash\_aqueous} --- uses 2-methyltetrahydrofuran in the
  organic phase with aqueous backwashing
\end{enumerate}
These processes are primarily workup and purification operations where MeTHF
serves as the organic solvent.
\end{quote}

\subsection{Expert assessment}

Two observations qualify this example. First, the two agents answer at different
levels of description: the lexical agent names synthesis steps in the report's own
vocabulary (S2, S3) with quantitative conditions, whereas the domain agent names
canonical unit operations. Both are legitimate readings of ``process step'', and a
router that combined them would return the resolved set \emph{and} the conditions.
Second, the domain agent's answer inherits its entity names from extraction, so
identifiers such as \texttt{dgf03\_037\_phase\_separation} are machine-generated rather
than terms a chemist would use; improving primary-name selection during
canonicalization would make these answers more readable without changing the
retrieval.

\end{document}